# A Hierarchical Graphical Model for Record Linkage

**Pradeep Ravikumar**
Center for Automated
Learning and Discovery,
School of Computer Science,
Carnegie Mellon University
pradeepr@cs.cmu.edu

**William W. Cohen**
Center for Automated
Learning and Discovery,
School of Computer Science,
Carnegie Mellon University
wcohen@cs.cmu.edu

## Abstract

The task of matching co-referent records is known among other names as record linkage. For large record-linkage problems, often there is little or no labeled data available, but unlabeled data shows a reasonably clear structure. For such problems, unsupervised or semi-supervised methods are preferable to supervised methods. In this paper, we describe a hierarchical graphical model framework for the record-linkage problem in an unsupervised setting. In addition to proposing new methods, we also cast existing unsupervised probabilistic record-linkage methods in this framework. Some of the techniques we propose to minimize overfitting in the above model are of interest in the general graphical model setting. We describe a method for incorporating monotonicity constraints in a graphical model. We also outline a bootstrapping approach of using "single-field" classifiers to noisily label latent variables in a hierarchical model. Experimental results show that our proposed unsupervised methods perform quite competitively even with fully supervised record-linkage methods.

## 1 Introduction

Databases frequently contain multiple records that refer to the same entity, but are not identical. The task of matching such co-referent records has been explored by a number of communities, including statistics, databases, and artificial intelligence. Each community has formulated the problem differently, and different techniques have been proposed.

In the database community, some work on record matching has been based on knowledge-intensive approaches [7, 6, 13]. More recently, the use of string-edit distances as a general-purpose record matching scheme was proposed by Monge and Elkan [10, 9], and in previous work [2, 3], we developed a toolkit of various string-distance based methods for matching entity-names. The AI community has focused on applying supervised learning to the record-linkage task — for learning the parameters of string-edit distance metrics [14, 1] and combining the results of different distance functions [15, 4, 1]. More recently, probabilistic object identification methods have been adapted to matching tasks [12]. In statistics, a long line of research has been conducted in *probabilistic record linkage*, largely based on the seminal paper by Fellegi and Sunter [5].

In this paper, we follow the Fellegi-Sunter approach of treating the record-linkage problem as a classification task, where the basic goal is to classify record-pairs as matching or non-matching. Many record-linkage problems are quite large, such as the matching of individuals and/or families between samples and censuses, *e.g.*, in the evaluation of the coverage of the U.S. decennial census. Often for such large problems, there is little or no labeled data available, but unlabeled data shows reasonably clear structure. For such problems, unsupervised or semi-supervised methods are preferable to supervised methods.

In this paper, we describe a hierarchical graphical model framework for approaching this problem. In addition to proposing new methods, we also cast existing unsupervised probabilistic record-linkage methods in the framework. The proposed graphical model has $(k+1)$ latent variables for records with $k$ fields, and hence fitting it to the data with minimal overfitting is a non-trivial task. We outline approaches to deal with this estimation problem in Section 4, some of which could also be utilized in more general graphical model applications. We address the problem of incorporating *monotonicity* constraints into a graphical model, which should be helpful in reducing overfitting in complex generative models where such constraints exist. We also outline a bootstrapping approach of using "single-field" classifiers to assign noisy labels to the latent variables in a hierarchical model. Results show that this enables us to capture constraints in the multi-field-record data more effectively. We also note that the proposed hierarchical model could be used to address the general problem of fitting a graphical model to continuous data (Section 7).

Experimental results show that our proposed unsupervised methods are competitive with fully supervised record-linkage methods.

## 2 Preliminaries

Given two lists of records, $A$ and $B$, we look at the the task of detecting the *matching* record-pairs $(a, b) \in A \times B$. A *record* is basically a vector of fields, *e.g.*, Figure 1. Thus, a record-pair is essentially a vector of field-pairs. More generally, we can represent a record-pair $(a, b)$ as a vector of features, often called a *comparison* vector: $f(a, b) = f_1(a, b), ..., f_k(a, b)$, where $f_1, \ldots, f_k$ are the features.

Unless specified otherwise, we will consider the record-pair feature vector to be a vector of *distances*, one for each field-pair. If there are $k$ fields, we denote the feature vector by $\mathbf{f}$, where $f_i$ is the *distance feature* for the $i$th field.

The record-linkage problem is the classification task of assigning the record-pair feature vectors to a label "matching" or "non-matching". Denote the match-class by a binary variable $M$, where $M = 0$ indicates a non-match and $M = 1$ indicates a match. The goal of probabilistic record-linkage is to formulate a probabilistic model for the match-class $M$ and the feature vector $\mathbf{f}$, and use the same to estimate the probability of the match class given the record-pair feature vector, $P(M|\mathbf{f})$. In an unsupervised setting, this amounts to estimating a generative model for $(\mathbf{f}, M)$.

## 3 Graphical Models for existing Record-Linkage Methods

Existing unsupervised methods for probabilistic record-linkage use a generative model for the record-pair feature vector with a single latent match-class variable, as shown in Figure 2. Note that the generative model is for the record-pair feature vector rather than the record-pair itself. Some other generative models for classification such as Naive-Bayes and Tree-Augmented Naive Bayes form special cases of Figure 2.

The predominant problem with the graphical model in Figure 2 is that the $f_i$ feature values are continuous, which precludes the normal multinomial probability model for a Bayesian network. There are two basic approaches to deal with continuous variables in a graphical model: we can restrict to specific families of parametric distributions *e.g.*, Gaussian mixture models, or we can discretize the variables and learn the model over the discrete domain.

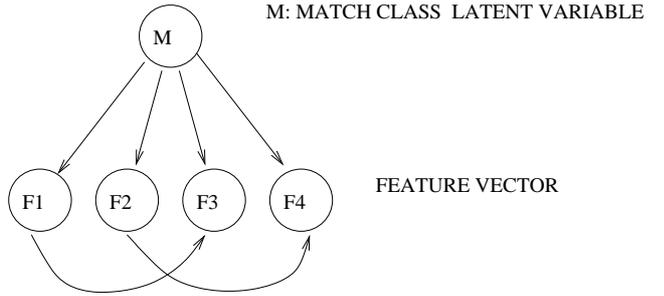

Figure 2: Model with a single latent match-class

**Discretization**: The $\mathbf{f}$ feature vector is discretized to a discrete-valued vector $\vec{w}$. For example, one way of discretizing into binary values is:

$$w_i = \begin{cases} 1 & \text{if } f_i > \theta_i \\ 0 & \text{if } f_i \leq \theta_i \end{cases} \quad (1)$$

One could then fit a multinomial probability model to the graphical model in Figure 2, as the observation values for the bottom layer are now discrete. This approach of binarization of the distance-features has been adopted by Winkler et al [16, 17] and is one of the baseline methods we compare our model to.

While the above approach enables using the efficient estimation and inference machinery of discrete graphical models, it has the problem that discretization into a small number of values leads to a poor approximation of the continuous distribution. However, if we increase the number of discrete values $d$, there is a potential explosion in the number of parameters of the model. In the graphical model of Figure 2, if the average number of parents for a node is $q$, and each node has $d$ values, then the number of multinomial parameters to estimate for $k$ such nodes is $O(kd^q)$. This might cause standard estimation methods to overfit the data.

**Using specific parametric families**: Instead of using the discrete-valued multinomial distribution for the variables, we could use other parametric families which allow continuous values, and for which there exists an efficient machinery for estimation and inference, such as the Gaussian distribution.

Gaussian distributions have the added advantage that a mixture of $m$ Gaussians can model any probability distribution to arbitrary accuracy, provided $m$ is large enough [8]. This suggests the following semi-supervised approach: we cluster the unlabeled feature-vectors using a Gaussian mixture model. We then use the limited labeled training data to label the clusters. The most frequently occurring label in the labeled feature vectors of a cluster is the label assigned to that cluster. This approach is another of the baselines we compare our model to.

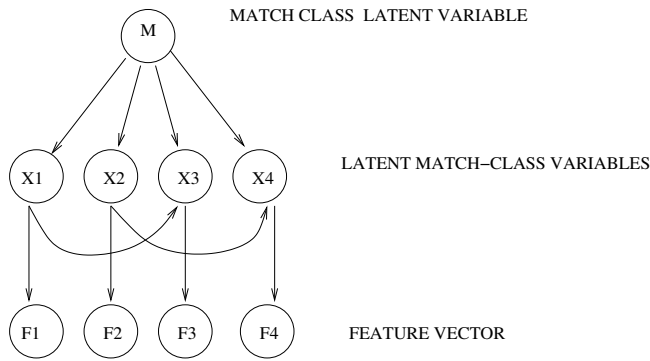

Figure 1: Records and Record Pairs

Figure 3: Hierarchical generative model for Record Linkage

**Supervised Learning**: One could also train the graphical model in Figure 2 in a supervised learning setting, if adequate labeled training data is available. As in [15, 4, 1], we could train, for example, a binary SVM classifier to predict the match-class given the continuous-valued feature-vector. Since this is a fully supervised method, it is not applicable when little or no labeled data is available.

## 4 A hierarchical graphical model for record-linkage

The graphical model for record linkage in Figure 2 can be generalized to a hierarchical three-layer model as shown in Figure 3. The bottom layer **f** in this model is a *feature-vector* layer as before, where each node in this layer corresponds to a distance-feature $f_i$. The difference in this model, when compared to Figure 2, is the set of binary latent variables $x_i$ for each distance feature $f_i$. The match-class latent variable $M$ in turn depends on these intermediate latent variables. Thus, we have a hierarchical mixture model with $(k+1)$ latent variables and $k$ dimensional observed data, given $k$ fields.

**Rationale for the hierarchical model**: While the intermediate latent variables as described above are operationally free to take any value, superimposing a certain semantic interpretation upon the latent variables shall give an intuition for the hierarchical model, as well as allow us to constrain the model in order to make the estimation of the structure and parameters easier.

Specifically, one could interpret the binary-valued middle layer **x** nodes in Figure 3 as latent match variables for each field. Thus, each node $x_i$ in the middle layer corresponds to the match-class of a single field-pair distance feature $f_i$. The top node in Figure 3 is the record-match class latent variable, which gives the match class of the entire record-pair, and which depends on the latent match class variables $x_i$ of the individual fields.

Thus, $P(x_i|f_i)$ gives the error model for field $i$. Assuming an independence of error models, there can be dependencies among nodes only in the middle layer, and all the bottom layer nodes are independent conditional on their latent-classes in the middle layer. This captures a natural intuition, since when we talk about dependencies between field-pairs, we imply a dependency between the match-classes of the field-pairs rather than the particular values themselves. Thus, when we say that the address field-pair is dependent on the name field-pair, we intuitively imply only that an address-match is dependent on a name-match. This is what is captured by the above graphical model.

As we model dependencies only between nodes in the middle layer, which are binary valued, we do not have to estimate as many parameters as a discretization of the model in Figure 2. Thus, assuming the average number of parents of a node in the middle layer is $q$, the number of multinomial parameters for any node in the middle layer is $O(2^q)$ and for all $k$ nodes is $O(k2^q)$. This is as opposed to $O(kd^q)$ if we directly modeled dependencies in Figure 2 after discretization as in the previous section.

However, the number of multinomial parameters in the graphical model of Figure 3 is still quite large, which is not surprising given that it is a hierarchical latent variable model with $(k+1)$ latent variables and $k$ observation variables. Hence, normal estimation techniques like EM would overfit the data. The remainder of the paper describes assumptions and constraints that reduce overfitting substantially: in fact, the final model does overfit, but performance is still very good and is competitive with supervised approaches.

## 5 Probability Estimation in the general model

Estimating the probabilities in the graphical model of Figure 3 via structural EM without any constraints is expensive with respect to both computation and generalization-error. Hence, we impose three types of constraints on the model before performing structural EM on the same. Our experiments thus follow the procedure below:

- We discretize each node in the bottom layer $f_i$ to have $d$ values. This does not cause a blowup of parameters since each bottom-layer node $f_i$ has only one parent $x_i$, and thus the number of multinomial parameters for each node $f_i$ is just $O(d)$.

- We then impose further constraints on the model as described in this section.

- Given the constraints, we then use structural EM to estimate the structure of the dependencies in the middle layer, and the parameters of the entire model.

### 5.1 Semantic Constraints

As described in Section 4, we can impose a semantic interpretation on the latent variables of the hierarchical graphical model: the middle layer **x** nodes as latent match-classes for the individual fields, and the top node $M$ as the match-class for the entire record-pair. This gives the constraint:

$$P(M=1|\mathbf{x}) = \begin{cases} 1 & \text{if } \mathbf{x} \text{ eq } \vec{1} \\ 0 & \text{otherwise} \end{cases} \quad (2)$$

The constraint in Equation 2 follows from the intuitive expectation that for a matching-record pair, the individual field-pairs would also be matches. Note that this does not assume that the field-pairs are error-free. In fact, given that the $x_i$ nodes represent the *true* match class for each individual field-pair $i$, $P(x_i|f_i)$ captures the error model for field $i$.

The match-class probability conditional on the observations can be estimated from Figure 3 as:

$$P(M=1|\mathbf{f}) = \sum_{\mathbf{x}} P(M=1, \mathbf{x}|\mathbf{f}) \quad (3)$$

As the overall match-class latent variable $M$ at the root is independent of the feature vector **f** of the bottom layer, given the latent match-classes **x**, we have

$$\sum_{\mathbf{x}} P(M=1, \mathbf{x}|\mathbf{f}) = \sum_{\mathbf{x}} P(M=1|\mathbf{x})P(\mathbf{x}|\mathbf{f}) \quad (4)$$

From Equations 3 and 4, and the constraint in Equation 2 it follows:

$$\begin{aligned} P(M=1|\mathbf{f}) &= \sum_{\mathbf{x}} P(M|\mathbf{x})P(\mathbf{x}|\mathbf{f}) \\ &= P(M=1|\mathbf{x}=\vec{1})P(\mathbf{x}=\vec{1}|\mathbf{f}) + 0 \\ &= P(\mathbf{x}=\vec{1}|\mathbf{f}) \end{aligned}$$

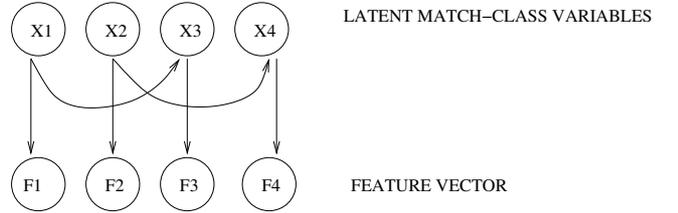

Figure 4: Two Layer Model

Thus, it follows that the dependencies between the latent variable $M$ and the latent variable layer **x** need not be modeled, and the graphical model in Figure 3 reduces to a non-hierarchical latent variable model with only the middle and bottom layers as shown in Figure 4. The record-linkage task thus reduces to a more manageable task of estimating $P(\mathbf{x}=\vec{1}|\mathbf{f})$ after learning the structure and parameters of the model in Figure 4.

### 5.2 Monotonicity Constraints

Given that the parent's label is "match", the probability of a node also having the label "match" is greater than having the label "non-match". The multinomial model of Figure 4 is still prone to overfitting as it does not capture such monotonicity constraints.

Specifically, if we discretize $f_i \in [0,1]$, with a higher discrete value indicating a higher degree of "match", we would like to enforce the monotonicity constraint:

$$j_1 \geq j_2 \Rightarrow P(f_i=j_1|x_i=1) \geq P(f_i=j_2|x_i=1)$$

We can capture these constraints by a simple modification to the multinomial model. Consider the standard multinomial model with $Pr(\mathbf{x}) = \prod_i p_i^{n_i}$ where $p_i$ are

the multinomial parameters and $n_i$ are the counts. We want to enforce additional constraints:

$$p_i \geq p_j \text{ if } i > j \quad (5)$$

We can do this by requiring $p_i = p_{i-1} + \Delta_i$ and constraining $\Delta_i$ to be positive. This leads to the reparametrization:

$$p_1 = \Delta_1, p_2 = \Delta_1 + \Delta_2, \ldots, p_k = \sum_{j=1}^{k} \Delta_j$$

$$\Delta_i \geq 0 \text{ for } i = 1, \ldots, k$$

The constraint in Equation 5 follows.

**Estimation of the $\Delta$ parameters**

The parameters $p_i$ of the standard multinomial model $Pr(\mathbf{x}) = \prod_i p_i^{n_i}$, when estimated via maximum likelihood, are given by $p_i = \frac{n_i}{\sum_i n_i}$.

We want to estimate the $\Delta_i$ parameters by maximizing the likelihood $Pr(\mathbf{x}) = \prod_i (\sum_{j=1}^{i} \Delta_j)^{n_i}$ under the constraints $\Delta_i \geq 0, \sum_i (\sum_{j=1}^{i} \Delta_j) = 1$. We cast this as an unconstrained optimization problem by using a Lagrange multiplier for the equality constraint, and using a barrier function for the inequality constraints. The barrier function attains value 0 if any $\Delta_i < 0$ and attains value 1 otherwise. In other words, we want the barrier function to be a step function at zero, but would also like it to be continuous and smooth to facilitate an easy optimization of the objective function. We use the sigmoid function $\sigma(x) = \frac{1}{1+e^{-ax}}$ for such a barrier function. In our experiments, we used $a = 20$. The optimization function then becomes:

$$\prod_i (\sum_{j=1}^{i} \Delta_j)^{n_i} \sigma(\Delta_i) - \lambda[\sum_i (\sum_{j=1}^{i} \Delta_j) - 1]$$

where $\lambda$ is a Lagrange multiplier.

On setting the derivative of the above function to zero, we get the following fixed point equations:

$$p_i = \frac{n_i - g_i p_i}{N - \sum_j g_j p_j} \quad (6)$$

where $N = \sum_i n_i$, $p_i = \sum_{j=1}^{i} \Delta_j$, $g_i = a(\sigma(\Delta_i) - \sigma(\Delta_{i+1}))$ for $i < k$ and $g_k = a(\sigma(\Delta_k) - 1)$.

The optimal estimates are thus given by the following iterative updates:

$$\hat{\Delta}_i^{(t)} = \hat{p}_i^{(t)} - \hat{p}_{i-1}^{(t)}$$
$$g_i^{(t)} = a(\sigma(\hat{\Delta}_i^{(t)}) - \sigma(\hat{\Delta}_{i+1}^{(t)}))$$
$$\hat{p}_i^{(t+1)} = \frac{n_i - g_i^{(t)} \hat{p}_i^{(t)}}{N - \sum_j g_j^{(t)} \hat{p}_j^{(t)}}$$

The convergence rate is quite fast empirically, averaging around 20 iterations for our datasets.

### 5.3 Bootstrapping with noisy labels

In this approach, we take a "single-field" classifier, which predicts the class-label $x_i$ given a single field-pair $f_i$, and use its output to noisily label the latent match-class variable $x_i$ given the observed field-pair feature $f_i$. Thus, in Figure 4, we have noisy labels for the latent match-class variables in the middle layer, and observed values for the distance feature variables in the bottom layer. We can thus learn the structure and parameters of the model as if in a completely supervised setting. In other words, we are able to bootstrap the model by allowing it to combine the outputs of $k$ single-field classifiers for each of the k fields.

Our experiments show that while training the entire model in this way, using noisy labels, performs better than using plain structural EM, there seems to be a problem of overfitting the noise in the labels. Thus, by labeling only a part of the unlabeled data using the noisy-label approach above, and performing EM on both the unlabeled and the noisily labeled data, as in [11], we are able to get even better results. In our experiments, we use a classifier based on the SoftTFIDF distance-metric[2].

## 6 Results

We have used two datasets to evaluate and compare the above methods, both of which have labeled records consisting of many correlated fields. The "census" dataset is a synthetic, census-like dataset containing 841 records, from which only textual fields were used (last name, first name, middle initial, house number, and street). The "restaurant" dataset contains 864 restaurant names and addresses with 112 duplicates, the fields being restaurant name, street address, city and cuisine.

Since it is not computationally practical to consider all pairs of records, we use a "blocking" method that outputs a smaller set of candidate pairs. For the moderate-size test sets considered here, we consider all pairs that share some character 4-gram. This 4-gram blocker finds an average of 99% of the correct pairs.

To evaluate a method on a dataset, we ranked all candidate pairs from the appropriate grouping algorithm by the posterior probability of the match-class given the observations. Following our earlier work [3], we computed the *non-interpolated average precision of this ranking*, the maximum *F1 score* of the ranking, and also *interpolated precision* at the eleven recall lev-els 0.0, 0.1, ..., 0.9, 1.0. *Precision* of a ranking containing $N$ pairs for a task with $m$ correct matches at a position $i$ is the fraction of pairs ranked before position $i$ that are correct, *i.e.*, $\frac{c(i)}{i}$ where $c(i)$ is the number of correct pairs ranked before position $i$. *Recall* at a position $i$ is the fraction of correct pairs ranked before position $i$, *i.e*, $\frac{c(i)}{m}$. *F1* score at a position $i$ is the harmonic mean of recall and precision at that position, $\frac{2pr}{p+r}$.

The *non-interpolated average precision* is $\frac{1}{m}\sum_{r=1}^{N}\frac{c(i)\delta(i)}{i}$, where $\delta(i) = 1$ if the pair at rank $i$ is correct and 0 otherwise. *Interpolated precision* at recall $r$ is the $\max_i \frac{c(i)}{i}$, where the max is taken over all ranks $i$ such that $\frac{c(i)}{m} \geq r$.

### 6.1 Baseline Methods

We compare our hierarchical model and its modifications to various baseline methods in Table 1. The baseline methods are described in detail in Section 2. The *Winkler unsup* method refers to the unsupervised EM-based estimation of parameters in the 2 layer model of Figure 2 as in [16, 17]. *Winkler sup* refers to a supervised Maximum Likelihood Estimation of the above 2 layer model given fully labeled data. *Winkler semisup* refers to a semi-supervised EM-based estimation of parameters given *partially* labeled data (one-third of a dataset) as in [11]. The Gaussian Mixture Model, described in Section 2, is also a semi-supervised method. Our experiments used a mixture model of 6 Gaussians. We used the SoftTFIDF distance metric [2] for all field distances. For the binarization in the Winkler methods above, a threshold of 0.8 was used. In the case of supervised and semi-supervised methods, three-fold cross-validation was used to evaluate performance.

Figure 5 compares our proposed hierarchical graphical model (HGM) to the above methods. Note that the HGM is a completely unsupervised method requiring no labeled data, and hence the comparison against semi-supervised and supervised methods stacks the odds against it. As Figure 5 shows, the HGM clearly outperforms baseline unsupervised methods, and is competitive with even fully supervised and semi-supervised methods.

### 6.2 Comparisons of modifications to the HGM

#### 6.2.1 Semantic Constraints

From Table 1, we see that the hierarchical model, in the absence of semantic constraints on the latent variables, is not able to fit the data at all. However, the with the addition of semantic constraints, *i.e.* the model of Figure 4, the performance of the model rises well above other baseline unsupervised methods.

#### 6.2.2 Monotonicity Constraints

As in Section 5.2, we modify the likelihood function by appending a barrier function, in order to satisfy certain monotonicity constraints implicit in the graphical model. From the table 1, we see that incorporating monotonicity constraints raise the performance markedly for the restaurant dataset, and slightly for the census dataset.

#### 6.2.3 Bootstrapping

As the Table 1 shows, training the hierarchical model (Figure 4) by noisily labeling the latent match-class nodes using a "single-field classifier" as described in Section 5.3, leads to a great improvement in performance. We also observe that adding the monotonicity constraints in addition to the bootstrapping, does not lead to a further increase in performance. This indicates that the bootstrapping approach is also able to constrain the model monotonically for this dataset. But for other datasets, adding both methods could conceivably improve performance to a greater extent.

## 7 Conclusions

For large record-linkage problems, often there is little or no labeled data available, but unlabeled data has reasonably clear structure. For such problems, unsupervised or semi-supervised methods are preferable to supervised methods. We have described a hierarchical latent variable graphical model (Figure 3) to perform record-linkage in such an unsupervised setting. Existing generative record-linkage methods can also be cast as special cases of the above general model. The hierarchical model has $k + 1$ latent variables for $k$ observations, and hence fitting it to the data with minimal overfitting is a non-trivial task. We outline approaches to the estimation problem which are applicable even in a general graphical model setting. We address the problem of incorporating *monotonicity* constraints in a graphical model, which should be helpful in reducing overfitting in complex generative models where such constraints exist. We also outline a bootstrapping approach of using a single-field classifier to assign noisy labels to the latent variables in the hierarchical model, which as results show, enable us to capture constraints in the data more effectively.

We also note that the above hierarchical model, and the estimation methods therein, can be used to address the general problem of fitting a graphical model to *continuous* data. For this continuous variable prob-lem, fitting a Gaussian mixture model performs poorly on our datasets, and discretization does not allow full freedom in modeling dependencies between variables as it suffers from a combinatorial explosion in the number of parameters. The method described in our paper, of introducing a latent variable for each node, and modeling dependencies only between the latent classes reduces the dimensionality of the parameter space considerably. Also sometimes, as with the record-linkage case, it makes more intuitive sense to model dependencies only in a latent-match-class layer.

| Method | Restaurant | | Census | |
|---|---|---|---|---|
| | AvgPrec | MaxF1 | AvgPrec | MaxF1 |
| Winkler   semi-supervised | 0.900 | 0.900 | 0.667 | 0.785 |
| supervised | 0.902 | 0.904 | 0.679 | 0.784 |
| Winkler   unsupervised | 0.617 | 0.568 | 0.495 | 0.612 |
| Gaussian Mixture Model | 0.702 | 0.704 | 0.242 | 0.388 |
| Hierarchical Graphical Model | 0.102 | 0.106 | 0.101 | 0.116 |
| Semantically Constrained HGM | 0.786 | 0.820 | 0.727 | 0.758 |
| + Monotonic Constraints | 0.795 | 0.823 | 0.728 | 0.759 |
| + Bootstrap | 0.820 | 0.844 | 0.728 | 0.759 |
| + both | 0.820 | 0.844 | 0.728 | 0.759 |

Table 1: Average precision and MaxF1 values for the record-linkage methods

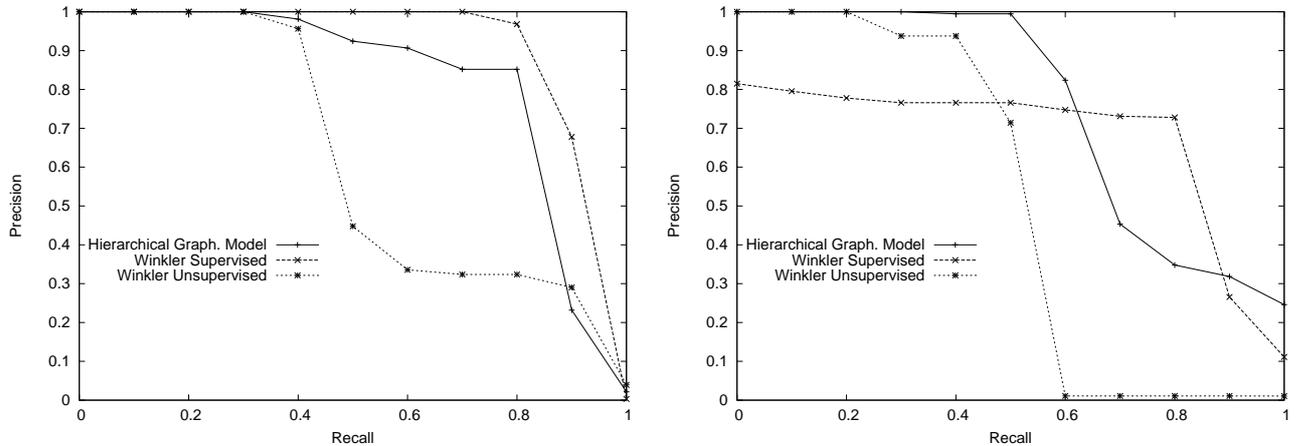

Figure 5: Comparison of the graphical models: Restaurant dataset to the left, Census dataset to the right

As Figure 5 shows, the unsupervised methods we propose are competitive with even fully supervised methods.